\documentclass[10pt,twocolumn,letterpaper]{article}

\usepackage{iccv}
\usepackage{times}
\usepackage{epsfig}
\usepackage{graphicx}
\usepackage{amsmath}
\usepackage{amssymb}

\usepackage[breaklinks=true,bookmarks=false]{hyperref}

\iccvfinalcopy 

\def\iccvPaperID{****} 

\ificcvfinal\fi

\newcommand{\bhline}{\noalign{\hrule height 1pt}}

\def\tbvoc{
\begin{table}[t]
\centering
\caption{Comparison of mean IoU on PASCAL VOC \texttt{test}~\cite{Everingham10,pascal-voc-2012}. The scores of previous works are cited from the literature~\cite{geng2021is}. Our model are evaluated by the test server.}
\vspace{1mm}
\label{tb:voc}
\begin{tabular}{lc}
\bhline
method & mean IoU \\ \hline
DANet~\cite{fu2019dual}  & 82.6     \\
DFN   ~\cite{yu2018learning} & 82.7     \\
PSPNet~\cite{zhao2017pyramid} & 82.6     \\
EncNet~\cite{zhang2018context} & 82.9     \\
SANet~\cite{zhong2020squeeze}  & 83.2     \\
APCNet~\cite{he2019adaptive} & 84.2     \\
CFNet ~\cite{zhang2019co} & 84.2     \\
SpyGR~\cite{li2020spatial}  & 84.2     \\
OCR  ~\cite{yuan2020object}  & 84.3     \\
DMNet~\cite{he2019dynamic}  & 84.4     \\
HamNet~\cite{geng2021is} & \textbf{85.9}     \\ \hline
Ours   & 83.9 \\ \bhline
\end{tabular}
\end{table}
}

\def\tbcontext{
\begin{table}[t]
\centering
\caption{Comparison of mean IoU on PASCAL Context \texttt{test}~\cite{mottaghi2014role}. The scores of previous works are cited from the literature~\cite{yuan2020object}.}
\vspace{1mm}
\label{tb:context}
\begin{tabular}{lc}
\bhline
Method     & mean IoU \\ \hline
PSPNet~\cite{zhao2017pyramid}     & 47.8 \\
SGR   ~\cite{liang2018symbolic}     & 50.8 \\
EncNet~\cite{zhang2018context}     & 51.7 \\
DANet ~\cite{fu2019dual}     & 52.6 \\
SpyGR ~\cite{li2020spatial}     & 52.8 \\
SANet ~\cite{zhong2020squeeze}     & 53.0 \\
EMANet~\cite{li2019expectation}     & 53.1 \\
CFNet ~\cite{zhang2019co}     & 54.0 \\
DMNet ~\cite{he2019dynamic}     & 54.4 \\
APCNet~\cite{he2019adaptive}     & 54.7 \\
OCR   ~\cite{yuan2020object}     & 54.8 \\
HamNet~\cite{geng2021is}     & 55.2 \\
OCR w/ HRNet~\cite{sun2019deep} backbone~\cite{yuan2020object} & 56.2 \\ \hline
Ours       & \textbf{56.4} \\ \bhline
\end{tabular}
\end{table}
}

\def\tbcityscapes{
\begin{table}[t]
\centering
\caption{Comparison of mean IoU on Cityscapes \texttt{test}~\cite{cordts2016cityscapes}. The scores of previous works are cited from the literature~\cite{yuan2020object}. Our model are evaluated by the test server.}
\vspace{1mm}
\label{tb:cityscapes}
\begin{tabular}{lc}
\bhline
Method    & mean IoU \\ \hline
DSSPN ~\cite{liang2018dynamic}     & 77.8 \\
SAC  ~\cite{zhang2017scale}     & 78.1 \\
Depth Seg~\cite{kong2018recurrent} & 78.2 \\
PSPNet  ~\cite{zhao2017pyramid}  & 78.4 \\
AAF     ~\cite{ke2018adaptive}  & 79.1 \\
CFNet  ~\cite{zhang2019co}  & 79.6 \\
PSANet  ~\cite{zhao2018psanet}  & 80.1 \\
PSPNet w/ coarse data ~\cite{zhao2017pyramid} & 81.2 \\
Deeplabv3 w/ coarse data ~\cite{chen2017rethinking} & 81.3 \\
ANNet   ~\cite{zhu2019asymmetric}  & 81.3 \\
BFP     ~\cite{ding2019boundary}  & 81.4 \\
CCNet   ~\cite{huang2019ccnet}   & 81.4 \\
OCR     ~\cite{yuan2020object}  & \textbf{81.8} \\ \hline
Ours      & 81.3 \\ \bhline
\end{tabular}
\end{table}
}

\def\tbade{
\begin{table}[t]
\centering
\caption{Comparison of mean IoU on ADE20K \texttt{valid}~\cite{zhou2017scene}. The scores of previous works are cited from the literature~\cite{yuan2020object}.}
\vspace{1mm}
\label{tb:ade}
\begin{tabular}{lc}
\bhline
Method & mean IoU  \\ \hline
SAC  ~\cite{zhang2017scale}  & 44.30 \\
PSPNet~\cite{zhao2017pyramid} & 43.29 \\
DSSPN ~\cite{liang2018dynamic} & 43.68 \\
PSANet~\cite{zhao2018psanet} & 43.77 \\
SGR   ~\cite{liang2018symbolic} & 44.32 \\
EncNet~\cite{zhang2018context} & 44.65 \\
GCU   ~\cite{li2018beyond} & 44.81 \\
CFNet ~\cite{zhang2019co} & 44.89 \\
CCNet ~\cite{huang2019ccnet} & 45.22 \\
ANNet ~\cite{zhu2019asymmetric} & 45.24 \\
OCR   ~\cite{yuan2020object} & 45.28 \\
APCNet~\cite{he2019adaptive} & \textbf{45.38} \\ \hline
Ours   & 44.59 \\ \bhline
\end{tabular}
\end{table}
}

\def\tbablation{
\begin{table}[t]
\centering
\caption{Ablation study on the depth of the hierarchy. In the table, None denotes the baseline. All models are evaluated without integration of the predictions.}
\vspace{1mm}
\label{tb:ablation}
\begin{tabular}{lc}
\bhline
Number of Hierarchy            & mean IoU \\ \hline
None                    & 49.5     \\
\{2\}                   & 54.0     \\
\{2, 4\}                & 54.5     \\
\{2, 4, 8\}             & 54.8     \\
\{2, 4, 8, 16\}         & \textbf{55.2}     \\
\{2, 4, 8, 16, 32\}     & 55.0     \\
\{2, 4, 8, 16, 32, 64\} & 55.0     \\ \bhline
\end{tabular}
\end{table}
}

\begin{document}

\title{Hierarchical Pyramid Representations for Semantic Segmentation}

\author{
\renewcommand{\thefootnote}{\fnsymbol{footnote}}
Hiroaki Aizawa${}^\text{1}$\thanks{Authors contributed equally} \thanks{Work done at Gifu University} \hspace{10mm}
Yukihiro Domae${}^\text{2}$$^*$ \hspace{10mm}
Kunihito Kato${}^\text{2}$\\
${}^\text{1}$Hiroshima University \hspace{10mm} ${}^\text{2}$Gifu Univeristy\\
\tt\small hiroaki-aizawa@hiroshima-u.ac.jp, domae@cv.info.gifu-u.ac.jp, kkato@gifu-u.ac.jp}

\maketitle
\ificcvfinal\thispagestyle{empty}\fi

\begin{abstract}
Understanding the context of complex and cluttered scenes is a challenging problem for semantic segmentation. However, it is difficult to model the context without prior and additional supervision because the scene's factors, such as the scale, shape, and appearance of objects, vary considerably in these scenes. To solve this, we propose to learn the structures of objects and the hierarchy among objects because context is based on these intrinsic properties. In this study, we design novel hierarchical, contextual, and multiscale pyramidal representations to capture the properties from an input image. Our key idea is the recursive segmentation in different hierarchical regions based on a predefined number of regions and the aggregation of the context in these regions. The aggregated contexts are used to predict the contextual relationship between the regions and partition the regions in the following hierarchical level. Finally, by constructing the pyramid representations from the recursively aggregated context, multiscale and hierarchical properties are attained. In the experiments, we confirmed that our proposed method achieves state-of-the-art performance in PASCAL Context.
\end{abstract}

\section{Introduction}
\label{sec:introduction}
Understanding the context is an open problem in semantic segmentation. The context is useful for recognizing objects in complex and cluttered scenes and objects with structure and articulation, such as humans and artifacts. This is because the context is provided by the individual objects in the scene and the background, and because the context of each object is formed from primitives parts of the object. Therefore, to understand the context, it is essential to learn the properties, such as the relationships between objects and the structures of the objects.

Modern approaches in semantic segmentation allow us to capture the long-range contextual information between objects in a scene. The long-range contexts are obtained by extending the receptive fields in convolutional neural networks~\cite{long2015fully,yu2016multi,peng2017large,zhang2017scale}, aggregating the multi-scale context using the pyramid representations~\cite{zhao2017pyramid,chen2017deeplab,chen2017rethinking,he2019adaptive,he2019dynamic}, and reasoning on graphs using a graph convolution~\cite{chen2019graph,li2018beyond,li2020spatial,hu2020class}.

\begin{figure}[t]
\centering
\includegraphics[width=1.0\linewidth]{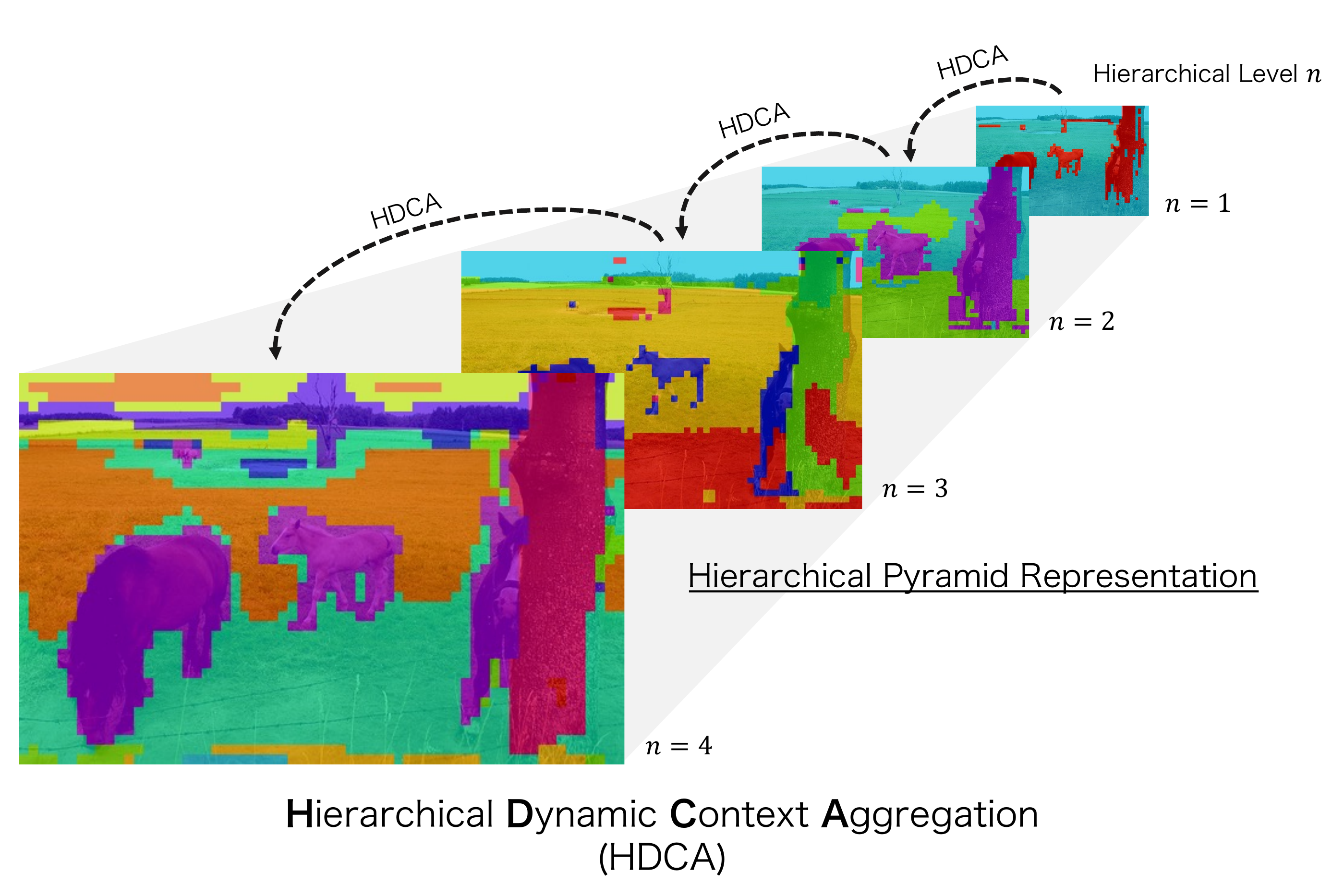}
\caption{Proposed concept. Our hierarchical dynamic context aggregation learns to aggregate hierarchical contexts from the recursively predicted regions with different receptive fields. We found that the hierarchical properties are essential for the understanding of the contexts and provide unsupervised hierarchical segmentation. More visual results are shown in Figure~\ref{fig:success}.}
\label{fig:concept}
\end{figure}

However, because these approaches focus on the aggregation of the global context in the image, they ignore the scene's hierarchy and object. The hierarchy is formed from the relationships in a set of primitives of the scene and object. Hence, the hierarchy helps the understanding of the context to recognize objects with complex structures in cluttered scenes. In human parsing, structural representation learning has been proposed by incorporating priors related to the hierarchy of the body parts into the models~\cite{wang2019learning,ji2019learning,yang2020renovating,wang2020hierarchical}. However, because these approaches are specific to the human body, it is difficult to apply them to other datasets (e.g., PASCAL Context~\cite{mottaghi2014role}).

In this paper, we propose to learn hierarchical, contextual, and multi-scale pyramid representations to capture the properties of the object's structure and the hierarchy in a scene. To achieve this, we design a novel mechanism we refer to as the \textbf{H}ierarchical \textbf{D}ynamic \textbf{C}ontext \textbf{A}ggregation~(HDCA). Compared with the conventional approaches for aggregating multi-scale contexts, as shown in Figure~\ref{fig:concept}, our key idea is to recursively divide the image of a scent into the various regions based on a predefined number of regions and aggregate the context in the regions. The aggregated contexts are used to predict the contextual relationship between the regions and partition the regions in the lower hierarchical level. Finally, the constructed pyramid representations from the recursively aggregated context has multi-scale and hierarchical properties. The proposed method allows us to learn the representations from semantic segmentation labels only without additional supervision or domain knowledge in complex and cluttered scenes. In the experiments, we confirmed that our proposed method achieves state-of-the-art performance in PASCAL Context.

We summarize our contributions as follows:
\begin{itemize}
    \item \textbf{Learning hierarchical and multi-scale contexts.} We propose a novel architecture equipping with hierarchical dynamic context aggregation for semantic segmentation. This allows us to aggregate the context depending on the object's region and shape and capture the contextual relationships between the regions. Moreover, this architecture can perform unsupervised hierarchical segmentation based on a predefined number of the regions.
    \item \textbf{Our proposed method achieves state-of-the-art performance in complex and cluttered scenes (e.g., PASCAL Context).} Moreover, the visualization results of unsupervised hierarchical segmentation showed that our method learns the hierarchical and multi-scale contexts, and the contextual relationships in the scene. 
\end{itemize}

\section{Related Work}
\label{sec:related}
\subsection{Context for Semantic Segmentation}
\label{ssec:related_context}
Semantic segmentation is a pixel-wise prediction task that assigns predefined classes to each pixel. The context is valid for this task. Specifically, the contextual relationships between pixels are utilized to distinguish objects with similar appearances and recognize distinct objects. The mainstream approaches extend the receptive fields by stacking convolution and pooling layers~\cite{long2015fully,yu2016multi,lin2017refinenet,peng2017large} to aggregate the global context. To further aggregate the context effectively, several approaches have been proposed based on the extraction of the multi-scale context~\cite{zhao2017pyramid,chen2017deeplab,chen2017rethinking,he2019adaptive,he2019dynamic}. In particular, pyramid scene parsing network~(PSPNet)~\cite{zhao2017pyramid} and DeepLab~\cite{chen2017deeplab} utilize the pooling with multiple sizes and dilated convolution with multiple dilation rates, respectively. As a result, these models allowed us to aggregate the context with various receptive fields from neighboring pixels. However, owing to the regular sampling size, the object's scale and shape were ignored, which leads to the confusion of classes with similar appearance and failure to extract dependencies of objects at a distance.

Our method aggregates the context irregularly, depending on the object's region. Similar to PSPNet, the aggregated contexts are merged as Pyramid features to obtain the multi-scale context.

\subsection{Attention-based Semantic Segmentation}
\label{ssec:related_attn}
Recently, attention-based semantic segmentation approaches~\cite{zhang2019acfnet,huang2019ccnet,fu2019dual,yuan2018ocnet,yuan2020object,li2019expectation,zhang2019co,zhao2018psanet,zhu2019asymmetric,geng2021is} have been studied to aggregate a region-wise context in a feature map. These works learn an attention map that describes the relationship in a scene based on the inner products of pairwise between pixels, and aggregate the context based on the attention map. Specifically, object-contextual representation~(OCR)~\cite{yuan2018ocnet,yuan2020object} learns the class-wise relation based on the elements in a feature map and prototypes of each class. However, because they only pay attention to contextual information within a class, they cannot capture the context across different scales or the relationships between classes.

Our method aggregates region-wise contexts to infer the contextual relationships between regions. To the best of our knowledge, OCR is the work closest to ours. The difference is that OCR learns intra-class relationships, while the proposed method acquires multi-scale contexts in a hierarchically segmented region.

\subsection{Graph-based Semantic Segmentation}
\label{ssec:related_graph}
In order to learn the dependencies of objects at a distance, semantic segmentation approaches using a graph convolution~\cite{kipf2017semi} have been proposed~\cite{chen2019graph,li2018beyond,li2020spatial,hu2020class,liang2018symbolic}. The graph convolution performs a convolution operation on a graph defined from a node and an edge and extracts the context on graph space. To construct the graph from a feature map, GCU~\cite{li2018beyond} constructs a graph with the regions in the feature maps as nodes by clustering the feature maps. Instead of clustering, GloRe~\cite{chen2019graph} reasons the regions to define the nodes and infers the contextual relationship between inferred regions using the graph convolution. However, graph-based approaches remain open problems in terms of the computational cost and the graph's construction from the image or feature map.

Instead of using a graph convolution, our method captures the relationship between the context in the segmented regions and feature maps based on their inner product of them.

\begin{figure*}[t]
\centering
\includegraphics[width=1.0\linewidth]{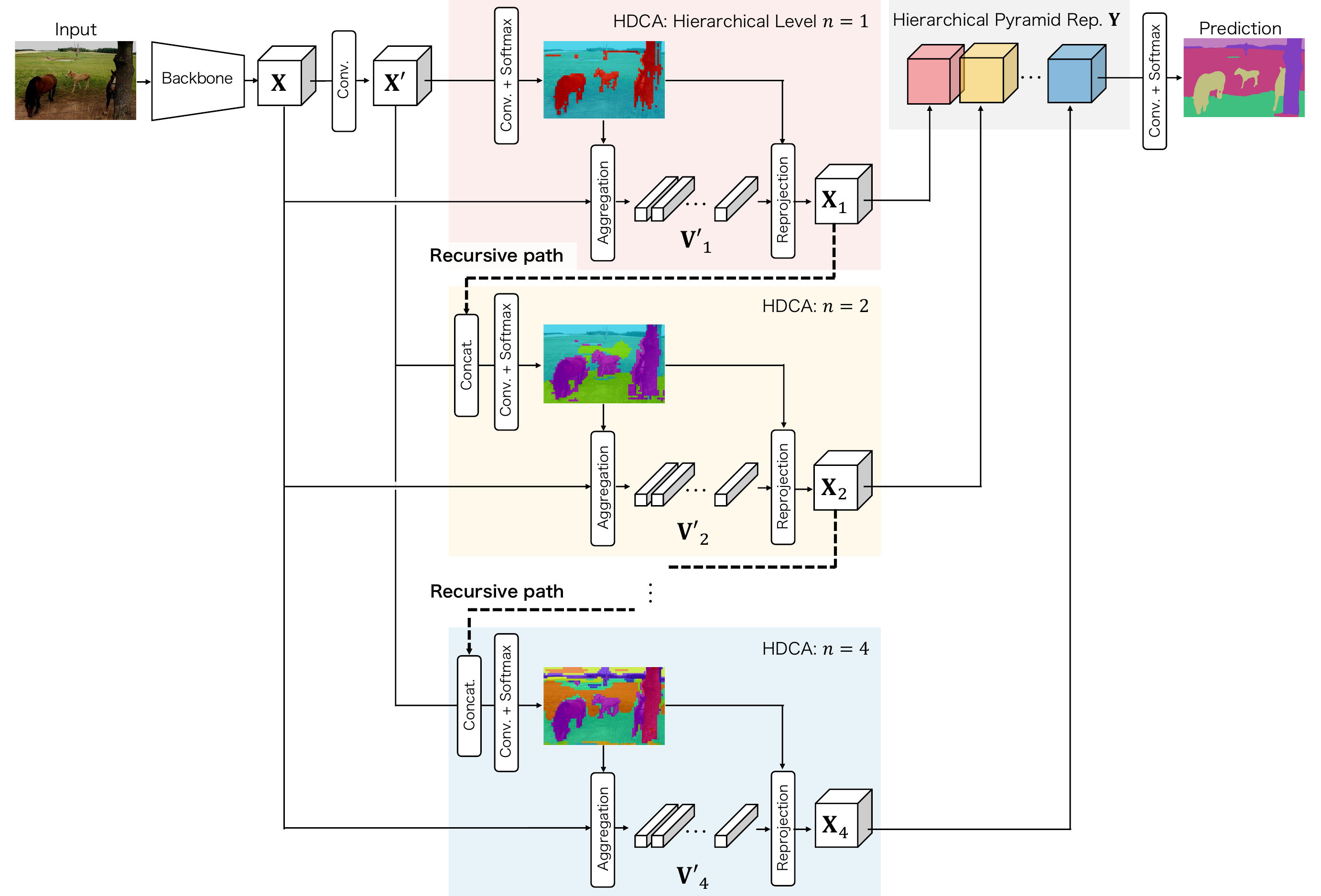}
\caption{Overview of our method with $N=4$ hierarchies. Our method consists of the backbone network and four HDCAs. The backbone network takes the RGB image as input and outputs the feature maps. Given the feature maps, each HDCA recursively divides them into the regions to aggregate the context vectors in the regions. The number of the segmented regions $\mathcal{S}_{4}$ at each hierarchical level is \{2, 4, 8, 16\}. The vectors are reprojected into feature map space. Finally, we build the pyramid representations from all hierarchical contexts. The proposed method allows us to learn the hierarchy and perform the unsupervised segmentation based on the predefined number of the regions without additional supervision.}
\label{fig:proposed}
\end{figure*}

\subsection{Hierarchy for Semantic Segmentation}
\label{ssec:related_hierarchy}
To represent a scene and an object hierarchically is essential for understanding the context. This is because the scene and the object consist of a set of primitives, and the relationship of the set provides the hierarchy. For example, a human's full-body consists of a head, and upper and lower bodies. We can further decompose the upper body into a torso, a right arm, and a left arm. These parts are dependent on each other and form a hierarchy with higher-level parts. Understanding the hierarchy helps to aggregate the context and recognize the fine-grained parts in an occluded scene. Therefore, researchers on human parsing pay attention to hierarchical reasoning~\cite{wang2019learning,ji2019learning,wang2020hierarchical}. Specifically, Wang~\textit{et al.} proposed a top-down/bottom-up inference for human parsing~\cite{wang2019learning}, and Ji~\textit{et al.} proposed to model the human's structure as a tree representation explicitly~\cite{ji2019learning}. However, they required domain knowledge of the human body and additional supervision of the hierarchy. Hence, it is difficult to apply them to complex and cluttered scene datasets~(e.g., PASCAL Context).

Several methods have also been proposed~\cite{liang2016semantic,liang2017interpretable,xiao2018unified} without the hierarchical labels in cases of more natural scenes . GraphLSTM~\cite{liang2016semantic} proposed by Liang~\textit{et al.} constructed a hierarchical graph based on superpixels' initial regions. However, these methods depend on the performance of the superpixels and tend to fail to construct the graph in complex and cluttered scenes.

Our proposed method allows us to learn the hierarchy without the additional supervisions and the priors. Moreover, it can apply to complex and cluttered scenes because it does not use superpixels.

\begin{figure*}[t]
\centering
\includegraphics[width=1.0\linewidth]{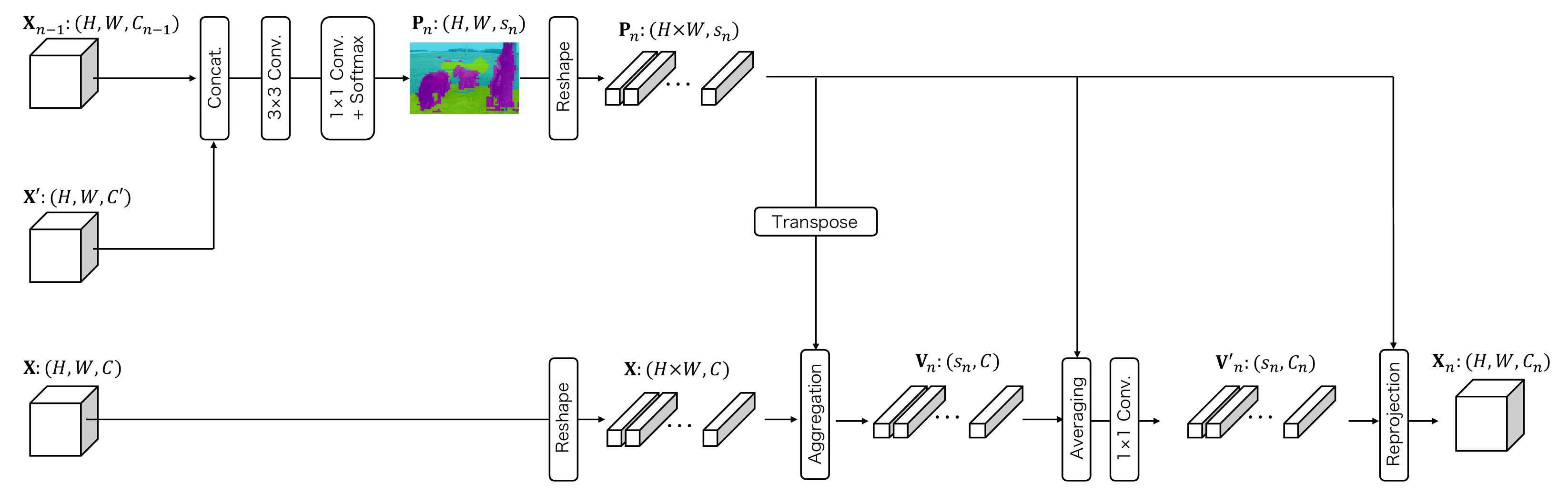}
\caption{Flowchart of the hierarchical dynamic context aggregation~(HDCA) process. The HDCA at $n$ hierarchical level takes the output $\mathbf{X}$ from the backbone network, $\mathbf{X'}$ reduced the dimensionality along with the channel direction from $\mathbf{X}$, and the output $\mathbf{X}_{n-1}$ from the previous hierarchy $n-1$. First, the HDCA predicts the regions $\mathbf{P}_n$. Next, it utilizes $\mathbf{P}_n$ to aggregate the contexts as region-wise context vectors. Finally, we reproject them into the feature map space. The aggregated contexts are used at the following hierarchy $n+1$. In the figure, the aggregation, averaging, and reprojection operations indicate equations~\eqref{eq:agg_v},~\eqref{eq:norm}, and~\eqref{eq:reproj}, respectively.}
\label{fig:hdca}
\end{figure*}

\section{Proposed Method}

\subsection{Overview}
Our goal is to learn to aggregate hierarchical, multi-scale contexts in a scene for semantic segmentation. An overview of our proposed model is shown in Figure~\ref{fig:proposed}. We assume that a depth of the hierarchy of the scene is $N$. The model takes the feature map $\mathbf{X}$ obtained from the backbone as input, and learns a hierarchical pyramid representation $\mathbf{Y} = \{\mathbf{X}_1, \mathbf{X}_2,...,\mathbf{X}_N \}$ by aggregating the context $N$ times recursively based on a predefined number of partitions $\mathcal{S}_{N} = \{s_n\}^{N}_{n=1}$ . To obtain a representation $\mathbf{X}_{n}$ depending on the interest region at each hierarchical level $n$, The model equips with a novel mechanism for aggregating the contexts referred to as hierarchical dynamic context aggregation (HDCA). As opposed to the conventional approaches for aggregating multi-scale contexts, HDCA utilizes the contexts $\mathbf{X}_{n-1}$ at a lower hierarchical level to recursively divide them in various regions, and to aggregates the context in these regions. The aggregated contexts are used to predict the contextual relationship between the regions. Finally, constructing the pyramid representations from the recursively aggregated context has multi-scale and hierarchical properties.

\subsection{Hierarchical Dynamic Context Aggregation}
The proposed model extracts hierarchical pyramid representations by coarse-to-fine prediction with different numbers of HDCAs to aggregate multi-scale contexts with scene hierarchy. For this purpose, Our model contains $N$ HDCAs, as shown in Figure~\ref{fig:proposed}.

To learn this representation, HDCA consists of three processes: inference of pooling regions, accumulation of contexts in the regions, and re-projection into feature maps. The flow in the hierarchy level $n$ is shown in Figure~\ref{fig:hdca}. HDCA in the hierarchy $n$ takes the output $\mathbf{X} \in \mathbb{R}^{H \times W \times C}$ from the backbone network, $\mathbf{X'} \in \mathbb{R}^{H \times W \times C'}$ reduced the dimensionality along with the channel direction from $\mathbf{X}$, and the output $\mathbf{X}_{n-1} \in \mathbb{R}^{H \times W \times C_{n-1}}$ from the previous hierarchy $n-1$. $H$ and $W$ are the height and width of the feature map, respectively. $C$, $C'$, and $C_n$ are channels of a feature map. Because there are no features $\mathbf{X}_{n-1}$ at the first level $n=1$, $\mathbf{X}$ and $\mathbf{X'}$ are input.

\subsubsection{Inference of Pooling Regions}
First, we concatenate $\mathbf{X'}$ and $\mathbf{X}_{n-1}$ along with the channel direction to infer the pooling regions for the context aggregation. We apply convolution with $3 \times 3$ kernel sizes and point-wise convolution with a predefined number of partitions $s_n$ as output channels. We then infer the regions for context aggregation by applying a softmax function. The probability map at this hierarchy stage is denoted by $\mathbf{P}_{n} \in [0,1]^{H \times W \times s_{n}}$.

We set a large number of partitions $s_n$ in the interest hierarchy than the number of partitions $s_{n-1}$ in the lower level. Therefore, we can utilize the lower level's hierarchical contexts $\mathbf{X}_{n-1}$ as cues to further divide the hierarchy into fine-grained semantic regions. This recursive segmentation provides the hierarchy.

\subsubsection{Accumulation of Contexts in the Regions}
Based on the segmented regions $\mathbf{P}_{n}$, the context within the regions is aggregated from $\mathbf{X}$. For this purpose, after reshaping $\mathbf{X}$ and $\mathbf{P}_{n}$, the contexts $\mathbf{V}_n \in \mathbb{R}^{s_n \times C}$ contained in each region are obtained by calculating the inner product. This inner product is defined as follows:
\begin{align}
\mathbf{V}_{n} = \mathbf{P}_{n}^{\top}\mathbf{X}.
\label{eq:agg_v}
\end{align}

For each region of $\mathbf{V}_n$ where the context for $s_n$ regions is aggregated, we calculate the average vector $\mathbf{v}_n' \in \mathbb{R}^{C_n}$ of features in the region. This is defined as follows, 
\begin{align}
\mathbf{v}_{n,i}'=\frac{\mathbf{v}_{n,i}}{\sum_{j=1}^{H \times W} \mathbf{P}_{i, j}},
\label{eq:norm}
\end{align}
where $i$ denotes the target region, and $\mathbf{v}_{n,i}$ is the context vector of that region at the $n$-th hierarchy. By doing this for all regions, $\mathbf{V}_n'=\{\mathbf{v}_{n,1}',...,\mathbf{v}_{n,s_n}'\} \subset \mathbb{R}^{C_n}$ is obtained. Subsequently, by performing $1\times1$ convolution operations for dimensionality reduction, the output contexts $\mathbf{V}' \in \mathbb{R}^{s_n \times C_{n}}$ are obtained.

\subsubsection{Reprojection into Feature Maps}
Finally, we reproject the average context vector into a feature map again using the aggregated contexts $\mathbf{V}_n'$ and regions $\mathbf{P}_{n}$. This re-projection is defined as follows:
\begin{align}
\mathbf{X}_{n}=\mathbf{P}_{n}\mathbf{V}_{n}'.
\label{eq:reproj}
\end{align}

Meanwhile, we reshape $\mathbf{X}_{n}$ to have the same width and height as those of the input feature map. The output of the HDCA at the $n$-th hierarchy is $\mathbf{X}_{n}$, which is used at the HDCA at the $n+1$-th hierarchy. To predict the final segmentation map, we use the pyramid representation, which concatenates the outputs of all $N$ levels of HDCA along with the channel direction.

\section{Evaluation}

\subsection{Evaluation Protocol}
\subsubsection{Implementation Details}
\noindent\textbf{Network Architecture.} Following previous studies~\cite{zhao2017pyramid,zhang2019co,yuan2020object}, we selected a backbone architecture and trained the proposed model. We employed ResNet101 as a backbone network, in which the last two pooling layers were removed and dilated convolutions were added. This backbone network was pre-trained on ImageNet and outputs a feature map that was downscaled to 1/8 of the input image size. In the decoder of the proposed method, we upsampled the final prediction until the output resolution by bilinear interpolation. The channels $C$, $C'$, and $\{C_n\}^N_{n=1}$ were 2048, 512, and 256, respectively.\\

\noindent\textbf{Optimization Method.} In training, we used stochastic gradient descent with momentum $=$ 0.9 and weight decay $=$ 0.0001. The learning rate was updated according to the polynomial learning rate policy $(1-(\frac{iter}{iter_{max}})^{0.9})$. We describe the initial learning rate and batch size in the following section (see Section~\ref{sssec:dataset}). In training, we used synchronized batch normalization instead of standard batch normalization in order to synchronize the mean and variance in batch normalization between the GPUs. Our model was implemented in PyTorch Encoding~\footnote{https://hangzhang.org/PyTorch-Encoding/}.\\

\noindent\textbf{Data Augmentation.} Following previous studies~\cite{zhao2017pyramid,zhang2019co,he2019dynamic,yuan2020object}, we performed a data augmentation in the training and test phases. In training, we performed random cropping, random scaling in the range [0.5,2], and random horizontal flipping with a probability of 0.5. We described the cropping size in the following section (see Section~\ref{sssec:dataset}). When evaluating on the Cityscapes dataset, we added random rotations in the range [-10, 10] and random Gaussian blurring. In the evaluation, we integrated the predictions applied with scaling and horizontal flip. For scaling, we set \{0.75, 1.0, ..., 2.25\} for Cityscapes, and \{0.5, 0.75, ..., 1.75\} for other datasets.

\subsubsection{Dataset}
\label{sssec:dataset}
\noindent\textbf{PASCAL VOC}~\cite{Everingham10,pascal-voc-2012} is the most competitive benchmark in semantic segmentation. It contains 20 classes and a background class, and most of the images contain a single object in the center. The training, validation, and test datasets contain 1,464, 1,449, and 1,456 images, respectively. Following previous studies~\cite{yuan2020object}, we added a semantic boundaries dataset~(SBD)~\cite{BharathICCV2011} re-annotated for semantic segmentation. The initial learning rate and batch size were 0.005 and 16, respectively. The random crop size was $512 \times 512$. \\

\noindent\textbf{PASCAL Context}~\cite{mottaghi2014role} is a dataset used for the study of the role of the context. It includes additional classes so that PASCAL VOC2010 can be re-annotated in detail. The total number of classes is 60. We trained and evaluated models on 59 classes, excluding the background class. The dataset contains 10,103 images, which are divided into 4,998 images for training and 5,105 images for test. The initial learning rate and batch size were 0.001 and 16, respectively. The random crop size was $480 \times 480$.\\

\noindent\textbf{Cityscapes}~\cite{cordts2016cityscapes} is a standard benchmark for understanding urban scenes, which contains 30 classes. Following prior studies~\cite{yuan2020object}, we used 19 classes for training and evaluation. The number of images for training, evaluation, and testing are 2,975, 500, and 1,525, respectively. The sizes of all images is $2,048 \times 1,024$.  In training, we did not use coarse label data but only used fine-annotated data. The initial learning rate and batch size were 0.004 and 8, respectively. The random crop size was $769 \times 769$.\\

\noindent\textbf{ADE20K}~\cite{zhou2017scene} is a large-scale dataset used for the understanding of general scenes. It consists of 150 categories of objects and stuff classes, which is a challenging dataset. The training, validation, and test datasets contain approximately 20K, 2K, and 3K images, respectively. The initial learning rate and batch size were 0.004 and 16, respectively. The random crop size for training was $480 \times 480$.

\tbvoc
\tbcontext
\tbcityscapes
\tbade

\subsection{Results on PASCAL VOC}
Table~\ref{tb:voc} shows the results on PASCAL VOC. This result shows that our method achieved competitive results with modern methods. Our method's performance was improved by 1\% compared with PSPNet; however, its performance was descreased by 2\% at most compared with other state-of-the-art methods. Because PASCAL VOC has a limited number of classes, and most of the images include a single object in the center, it is a simple dataset. Hence, although our method outperformed PSPNet, this result shows that the hierarchy's benefit on this dataset is small.

\begin{figure*}[t]
\centering
\includegraphics[width=1.0\linewidth]{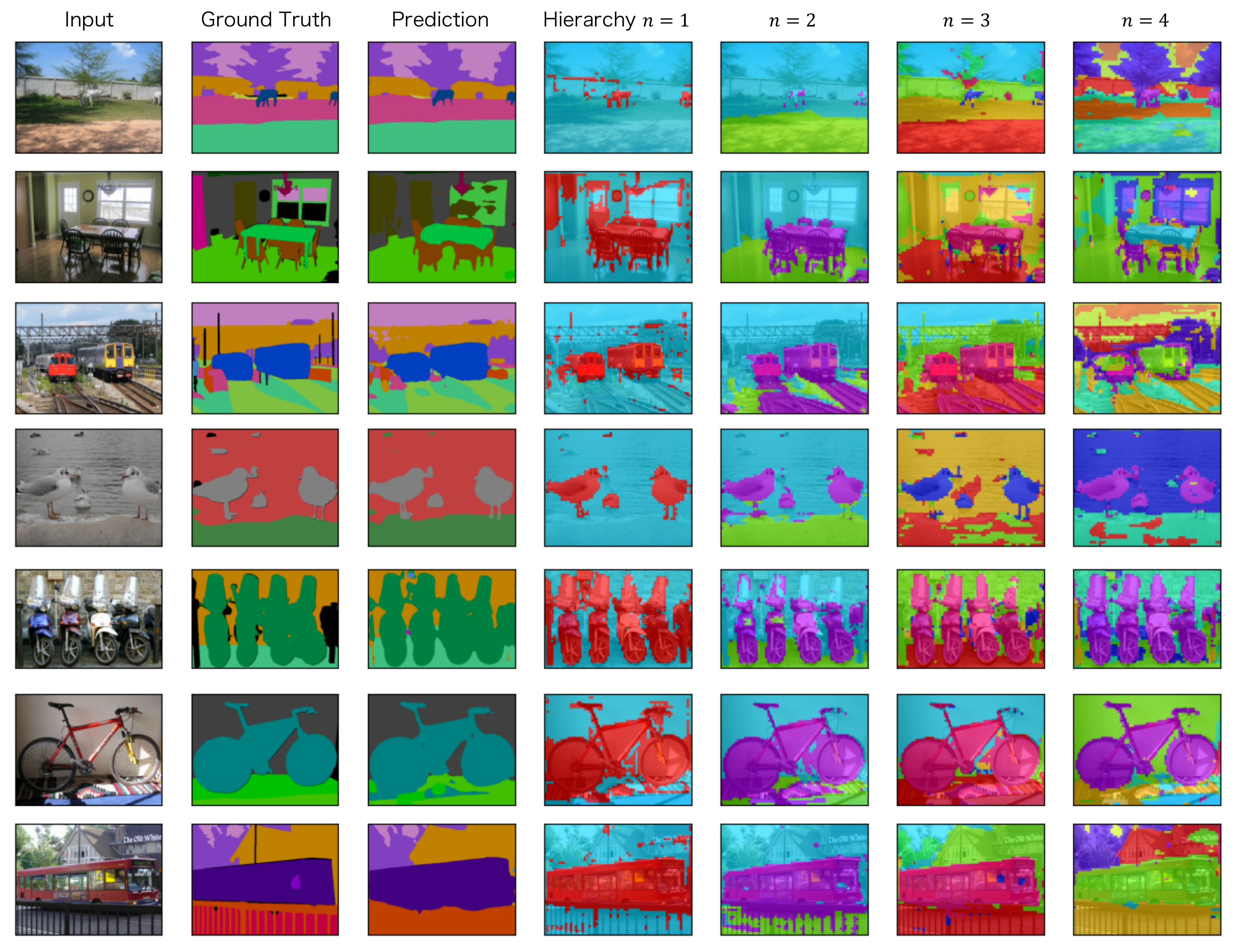}
\caption{Semantic segmentation results and visualization of the hierarchical segmentation for context aggregation on PASCAL Context~\texttt{test}. These results are obtained from our $\mathcal{S}_{4}$ model. From left to right: input image, ground truth, final prediction, and segmentation results at each hierarchical level. Note that the color in the hierarchical segmentation is consistent across hierarchies but does not indicate semantics since these colors are determined from the indices of the regions.}
\label{fig:success}
\end{figure*}

\subsection{Results on PASCAL Context}
Table~\ref{tb:context} is the result of the test data of PASCAL Context. This shows that our method outperforms the existing methods and achieves state-of-the-art performance. Because PASCAL Context has a more significant number of classes compared with the PASCAL VOC described above, understanding the context is essential for accurate segmentation. The results reveal that the hierarchy helps the understanding of the context and contributes to the improvement of performance.

\subsection{Results on Cityscapes}
To evaluate the effectiveness, we present the results of Cityscapes in Table~\ref{tb:cityscapes}. Surprisingly, despite our training without the extra labels, the performance is close to PSPNet and Deeplab v3 with it. Additionally, the result shows that competitive performance was achieved with DANet and OCR. These models introduce an auxiliary loss to stabilize the training and to guarantee for maintaining the spatial information. On the other hand, the proposed method learns only from the loss of the final prediction. Moreover, it does not use additional supervision to learn hierarchical representations. A technique that enforces the capturing of the hierarchy and the stabilization of the training will be the subject of our future work.

\subsection{Results on ADE20K}
We show the evaluation results on ADE20K on Table~\ref{tb:ade}. According to Table~\ref{tb:ade}, because the mean IoU of our methods was improved by 1.3\% comapred with PSPNet (which used pyramid representations from regular grid sampling), we confirmed the effectiveness of the hierarchy for aggregating the context. However, compared with other sophisticated approaches, it was down to 0.7\%. The variation of factors in a scene, such as the object and stuff categories and layout, is significant. We showed that designing and learning the hierarchy for understanding the context in such a scene is an open problem.

\begin{figure*}[t]
\centering
\includegraphics[width=1.0\linewidth]{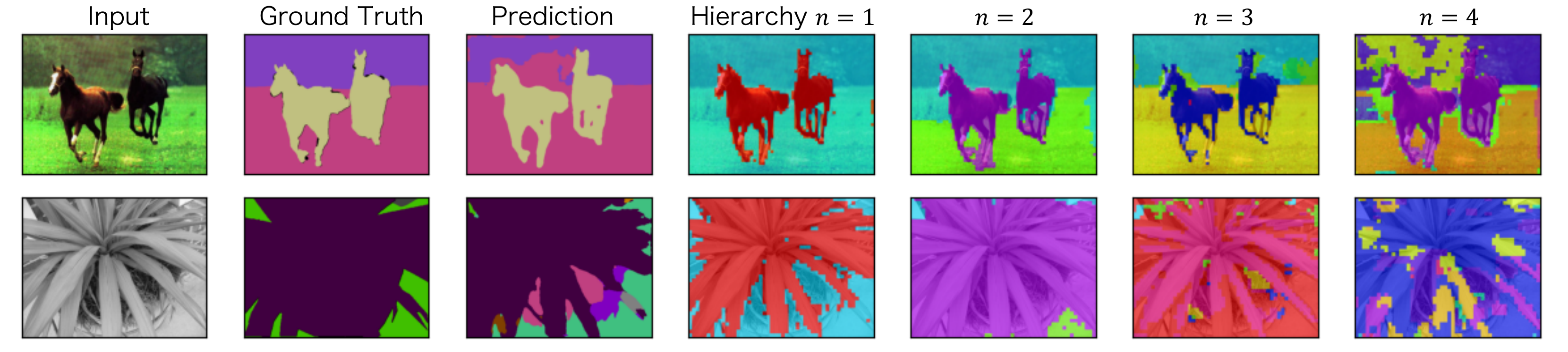}
\caption{Failure examples on PASCAL Context~\texttt{test}.}
\label{fig:failure}
\end{figure*}

\tbablation

\subsection{Ablation Study on the Depth of the Hierarchy}
We evaluated the correlation between the performance and the depth of the hierarchy based on the PASCAL Context dataset. We compared our method with conventional models without the hierarchy. We selected the fully convolutional architecture with a ResNet101 backbone as the baseline. Regarding the depth of the hierarchy, we designed the proposed models by dividing them into $\mathcal{S}_{1}=\{2\}$,
$\mathcal{S}_{2}=\{2,4\}$, $\mathcal{S}_{3}=\{2,4,8\}$, $\mathcal{S}_{4}=\{2,4,8,16\}$, $\mathcal{S}_{5}=\{2,4,8,16,32\}$, and $\mathcal{S}_{6}=\{2,4,8,16,32,64\}$ regions. All networks were trained from scratch. Obtained results are listed on Table~\ref{tb:ablation}. As shown in Table~\ref{tb:ablation}, the best hierarchy was $\mathcal{S}_{4}$ and the mean IoU was 55.2. Compared with the baseline, the proposed method of $\mathcal{S}_{4}$ was improved by 5.7\%. Interestingly, the mean IoU increased with the increase of the depth, and the peak was at $\mathcal{S}_{4}$.

The mean IoUs of $\mathcal{S}_{5}$ and $\mathcal{S}_{6}$ were almost the same as the $\mathcal{S}_{4}$ hierarchy. These results show that the number of predicted regions of $\mathcal{S}_{5}$ and $\mathcal{S}_{6}$ does not increase. This is because it is difficult to divide them into more regions compared with the number of categories in the input image without additional hierarchical supervision.

\subsection{Visualization on Hierarchical Segmentation}
To validate the learnt hierarchy, we visualized the hierarchical segmentation of $\mathcal{S}_{4}$ based on the PASCAL Context dataset. The results are shown in Figure~\ref{fig:success}. Figure~\ref{fig:success} shows the input image, ground truth, the final prediction, and hierarchical segmentation results of $\mathcal{S}_{4}=\{2,4,8,16\}$. Note that the color in the hierarchical segmentation does not indicate semantics since these colors are determined from the indices of the regions. As it can be observed in Figure~\ref{fig:success}, HDCA allows us to perform the hierarchical segmentation without any hierarchical labels. For example, as shown in Figure~\ref{fig:success}, HDCA of a first hierarchy divides an image into the foreground and the background. In a second hierarchy, a wall, a ground, and the foreground were divided, and the wall was further decomposed in a third hierarchy. Finally, the segmentation result at a fourth level was close to the final prediction. Figure~\ref{fig:success} indicates that HDCA can learn the hierarchical and multi-scale context because it aggregates the hierarchy-wise context from these regions and builds pyramid representations from the contexts.

Furthermore, the fourth and fifth rows in Figure~\ref{fig:success} shows that HDCA captures the dependencies of multiple objects assigned to the same category in the image. As it can be observed in these rows, the objects, such as birds and motorcycles, are divided into consistent regions across the hierarchy. Furthermore, as shown in the sixth and seventh rows in Figure~\ref{fig:success}, we confirmed that our method can construct the hierarchy even when the number of objects in the image is smaller than the predefined number of decomposition.

Finally, we show the failure case on Figure~\ref{fig:failure}. As shown in the failure example, HDCA tends to oversegment the region, and such representation makes the prediction more difficult. Hereafter, we will tackle the problem of the control of the number of regions and hierarchies according to the input image.

\section{Conclusion}

In this paper, we addressed the problem of learning the hierarchy for semantic segmentation. We started with the hypothesis that hierarchy was essential for the understanding of the context in a scene because objects and scenes were composed of primitives and proposed to learn hierarchical representations by recursively dividing them into regions and by aggregating the contexts within the regions. The proposed novel context aggregation method, Hierarchical Dynamic Context Aggregation, can perform hierarchical scene decomposition without additional supervision.

In the experiments, we validated the effectiveness of our idea from the hierarchical decomposition and segmentation performance perspectives. We found that the depth of the hierarchy was related to the performance and that the contextual dependencies across the hierarchy were effective for semantic segmentation. However, one of our work's limitations pretains to the design and learning of the hierarchy when the dataset includes a significant variation in the scene, such as the object's category, stuff, and layout. Our future study will be focused on this issue. We believe that a continuous study of hierarchical representation learning would demonstrate that the hierarchy is essential not only for semantic segmentation but also for the interpretability of deep neural networks and other computer vision tasks.

\section*{Acknowledgement}
This work is partially supported by Grants-in-aid for Promotion of Regional Industry-University-Government Collaboration from Cabinet Office, Japan.

{\small
\bibliographystyle{ieee_fullname}
\bibliography{egbib}

\begin{thebibliography}{10}\itemsep=-1pt

\bibitem{chen2017deeplab}
Liang-Chieh Chen, George Papandreou, Iasonas Kokkinos, Kevin Murphy, and Alan~L
  Yuille.
\newblock Deeplab: Semantic image segmentation with deep convolutional nets,
  atrous convolution, and fully connected crfs.
\newblock {\em IEEE transactions on pattern analysis and machine intelligence},
  40(4):834--848, 2017.

\bibitem{chen2017rethinking}
Liang-Chieh Chen, George Papandreou, Florian Schroff, and Hartwig Adam.
\newblock Rethinking atrous convolution for semantic image segmentation.
\newblock {\em arXiv preprint arXiv:1706.05587}, 2017.

\bibitem{chen2019graph}
Yunpeng Chen, Marcus Rohrbach, Zhicheng Yan, Yan Shuicheng, Jiashi Feng, and
  Yannis Kalantidis.
\newblock Graph-based global reasoning networks.
\newblock In {\em CVPR}, 2019.

\bibitem{cordts2016cityscapes}
Marius Cordts, Mohamed Omran, Sebastian Ramos, Timo Rehfeld, Markus Enzweiler,
  Rodrigo Benenson, Uwe Franke, Stefan Roth, and Bernt Schiele.
\newblock The cityscapes dataset for semantic urban scene understanding.
\newblock In {\em CVPR}, 2016.

\bibitem{ding2019boundary}
Henghui Ding, Xudong Jiang, Ai~Qun Liu, Nadia~Magnenat Thalmann, and Gang Wang.
\newblock Boundary-aware feature propagation for scene segmentation.
\newblock In {\em ICCV}, 2019.

\bibitem{pascal-voc-2012}
M. Everingham, L. Van~Gool, C.~K.~I. Williams, J. Winn, and A. Zisserman.
\newblock The {PASCAL} {V}isual {O}bject {C}lasses {C}hallenge 2012 {(VOC2012)}
  {R}esults.
\newblock
  http://www.pascal-network.org/challenges/VOC/voc2012/workshop/index.html.

\bibitem{Everingham10}
M. Everingham, L. Van~Gool, C.~K.~I. Williams, J. Winn, and A. Zisserman.
\newblock The pascal visual object classes (voc) challenge.
\newblock {\em International Journal of Computer Vision}, 88(2):303--338, June
  2010.

\bibitem{fu2019dual}
Jun Fu, Jing Liu, Haijie Tian, Yong Li, Yongjun Bao, Zhiwei Fang, and Hanqing
  Lu.
\newblock Dual attention network for scene segmentation.
\newblock In {\em CVPR}, 2019.

\bibitem{geng2021is}
Zhengyang Geng, Meng-Hao Guo, Hongxu Chen, Xia Li, Ke Wei, and Zhouchen Lin.
\newblock Is attention better than matrix decomposition?
\newblock In {\em ICLR}, 2021.

\bibitem{BharathICCV2011}
Bharath Hariharan, Pablo Arbelaez, Lubomir Bourdev, Subhransu Maji, and
  Jitendra Malik.
\newblock Semantic contours from inverse detectors.
\newblock In {\em ICCV}, 2011.

\bibitem{he2019dynamic}
Junjun He, Zhongying Deng, and Yu Qiao.
\newblock Dynamic multi-scale filters for semantic segmentation.
\newblock In {\em ICCV}, 2019.

\bibitem{he2019adaptive}
Junjun He, Zhongying Deng, Lei Zhou, Yali Wang, and Yu Qiao.
\newblock Adaptive pyramid context network for semantic segmentation.
\newblock In {\em CVPR}, 2019.

\bibitem{hu2020class}
Hanzhe Hu, Deyi Ji, Weihao Gan, Shuai Bai, Wei Wu, and Junjie Yan.
\newblock Class-wise dynamic graph convolution for semantic segmentation.
\newblock In {\em ECCV}, 2020.

\bibitem{huang2019ccnet}
Zilong Huang, Xinggang Wang, Lichao Huang, Chang Huang, Yunchao Wei, and Wenyu
  Liu.
\newblock Ccnet: Criss-cross attention for semantic segmentation.
\newblock In {\em ICCV}, 2019.

\bibitem{ji2019learning}
Ruyi Ji, Dawei Du, Libo Zhang, Longyin Wen, Yanjun Wu, Chen Zhao, Feiyue Huang,
  and Siwei Lyu.
\newblock Learning semantic neural tree for human parsing.
\newblock In {\em ECCV}, 2020.

\bibitem{ke2018adaptive}
Tsung-Wei Ke, Jyh-Jing Hwang, Ziwei Liu, and Stella~X Yu.
\newblock Adaptive affinity fields for semantic segmentation.
\newblock In {\em ECCV}, 2018.

\bibitem{kipf2017semi}
Thomas~N Kipf and Max Welling.
\newblock Semi-supervised classification with graph convolutional networks.
\newblock In {\em ICLR}, 2017.

\bibitem{kong2018recurrent}
Shu Kong and Charless~C Fowlkes.
\newblock Recurrent scene parsing with perspective understanding in the loop.
\newblock In {\em CVPR}, 2018.

\bibitem{li2020spatial}
Xia Li, Yibo Yang, Qijie Zhao, Tiancheng Shen, Zhouchen Lin, and Hong Liu.
\newblock Spatial pyramid based graph reasoning for semantic segmentation.
\newblock In {\em CVPR}, 2020.

\bibitem{li2019expectation}
Xia Li, Zhisheng Zhong, Jianlong Wu, Yibo Yang, Zhouchen Lin, and Hong Liu.
\newblock Expectation-maximization attention networks for semantic
  segmentation.
\newblock In {\em ICCV}, 2019.

\bibitem{li2018beyond}
Yin Li and Abhinav Gupta.
\newblock Beyond grids: Learning graph representations for visual recognition.
\newblock In {\em NeurIPS}, 2018.

\bibitem{liang2018symbolic}
Xiaodan Liang, Zhiting Hu, Hao Zhang, Liang Lin, and Eric~P Xing.
\newblock Symbolic graph reasoning meets convolutions.
\newblock In {\em NeurIPS}, 2018.

\bibitem{liang2017interpretable}
Xiaodan Liang, Liang Lin, Xiaohui Shen, Jiashi Feng, Shuicheng Yan, and Eric~P
  Xing.
\newblock Interpretable structure-evolving lstm.
\newblock In {\em CVPR}, 2017.

\bibitem{liang2016semantic}
Xiaodan Liang, Xiaohui Shen, Jiashi Feng, Liang Lin, and Shuicheng Yan.
\newblock Semantic object parsing with graph lstm.
\newblock In {\em ECCV}, 2016.

\bibitem{liang2018dynamic}
Xiaodan Liang, Hongfei Zhou, and Eric Xing.
\newblock Dynamic-structured semantic propagation network.
\newblock In {\em CVPR}, 2018.

\bibitem{lin2017refinenet}
Guosheng Lin, Anton Milan, Chunhua Shen, and Ian Reid.
\newblock Refinenet: Multi-path refinement networks for high-resolution
  semantic segmentation.
\newblock In {\em CVPR}, 2017.

\bibitem{long2015fully}
Jonathan Long, Evan Shelhamer, and Trevor Darrell.
\newblock Fully convolutional networks for semantic segmentation.
\newblock In {\em CVPR}, 2015.

\bibitem{mottaghi2014role}
Roozbeh Mottaghi, Xianjie Chen, Xiaobai Liu, Nam-Gyu Cho, Seong-Whan Lee, Sanja
  Fidler, Raquel Urtasun, and Alan Yuille.
\newblock The role of context for object detection and semantic segmentation in
  the wild.
\newblock In {\em CVPR}, 2014.

\bibitem{peng2017large}
Chao Peng, Xiangyu Zhang, Gang Yu, Guiming Luo, and Jian Sun.
\newblock Large kernel matters--improve semantic segmentation by global
  convolutional network.
\newblock In {\em CVPR}, 2017.

\bibitem{sun2019deep}
Ke Sun, Bin Xiao, Dong Liu, and Jingdong Wang.
\newblock Deep high-resolution representation learning for human pose
  estimation.
\newblock In {\em CVPR}, 2019.

\bibitem{wang2019learning}
Wenguan Wang, Zhijie Zhang, Siyuan Qi, Jianbing Shen, Yanwei Pang, and Ling
  Shao.
\newblock Learning compositional neural information fusion for human parsing.
\newblock In {\em ICCV}, 2019.

\bibitem{wang2020hierarchical}
Wenguan Wang, Hailong Zhu, Jifeng Dai, Yanwei Pang, Jianbing Shen, and Ling
  Shao.
\newblock Hierarchical human parsing with typed part-relation reasoning.
\newblock In {\em CVPR}, 2020.

\bibitem{xiao2018unified}
Tete Xiao, Yingcheng Liu, Bolei Zhou, Yuning Jiang, and Jian Sun.
\newblock Unified perceptual parsing for scene understanding.
\newblock In {\em ECCV}, 2018.

\bibitem{yang2020renovating}
Lu Yang, Qing Song, Zhihui Wang, Mengjie Hu, Chun Liu, Xueshi Xin, Wenhe Jia,
  and Songcen Xu.
\newblock Renovating parsing r-cnn for accurate multiple human parsing.
\newblock In {\em ECCV}, 2020.

\bibitem{yu2018learning}
Changqian Yu, Jingbo Wang, Chao Peng, Changxin Gao, Gang Yu, and Nong Sang.
\newblock Learning a discriminative feature network for semantic segmentation.
\newblock In {\em CVPR}, 2018.

\bibitem{yu2016multi}
Fisher Yu and Vladlen Koltun.
\newblock Multi-scale context aggregation by dilated convolutions.
\newblock In {\em ICLR}, 2016.

\bibitem{yuan2020object}
Yuhui Yuan, Xilin Chen, and Jingdong Wang.
\newblock Object-contextual representations for semantic segmentation.
\newblock In {\em ECCV}, 2020.

\bibitem{yuan2018ocnet}
Yuhui Yuan and Jingdong Wang.
\newblock Ocnet: Object context network for scene parsing.
\newblock {\em arXiv preprint arXiv:1809.00916}, 2018.

\bibitem{zhang2019acfnet}
Fan Zhang, Yanqin Chen, Zhihang Li, Zhibin Hong, Jingtuo Liu, Feifei Ma, Junyu
  Han, and Errui Ding.
\newblock Acfnet: Attentional class feature network for semantic segmentation.
\newblock In {\em ICCV}, 2019.

\bibitem{zhang2018context}
Hang Zhang, Kristin Dana, Jianping Shi, Zhongyue Zhang, Xiaogang Wang, Ambrish
  Tyagi, and Amit Agrawal.
\newblock Context encoding for semantic segmentation.
\newblock In {\em CVPR}, 2018.

\bibitem{zhang2019co}
Hang Zhang, Han Zhang, Chenguang Wang, and Junyuan Xie.
\newblock Co-occurrent features in semantic segmentation.
\newblock In {\em CVPR}, 2019.

\bibitem{zhang2017scale}
Rui Zhang, Sheng Tang, Yongdong Zhang, Jintao Li, and Shuicheng Yan.
\newblock Scale-adaptive convolutions for scene parsing.
\newblock In {\em ICCV}, 2017.

\bibitem{zhao2017pyramid}
Hengshuang Zhao, Jianping Shi, Xiaojuan Qi, Xiaogang Wang, and Jiaya Jia.
\newblock Pyramid scene parsing network.
\newblock In {\em CVPR}, 2017.

\bibitem{zhao2018psanet}
Hengshuang Zhao, Yi Zhang, Shu Liu, Jianping Shi, Chen~Change Loy, Dahua Lin,
  and Jiaya Jia.
\newblock Psanet: Point-wise spatial attention network for scene parsing.
\newblock In {\em ECCV}, 2018.

\bibitem{zhong2020squeeze}
Zilong Zhong, Zhong~Qiu Lin, Rene Bidart, Xiaodan Hu, Ibrahim~Ben Daya, Zhifeng
  Li, Wei-Shi Zheng, Jonathan Li, and Alexander Wong.
\newblock Squeeze-and-attention networks for semantic segmentation.
\newblock In {\em CVPR}, 2020.

\bibitem{zhou2017scene}
Bolei Zhou, Hang Zhao, Xavier Puig, Sanja Fidler, Adela Barriuso, and Antonio
  Torralba.
\newblock Scene parsing through ade20k dataset.
\newblock In {\em CVPR}, 2017.

\bibitem{zhu2019asymmetric}
Zhen Zhu, Mengde Xu, Song Bai, Tengteng Huang, and Xiang Bai.
\newblock Asymmetric non-local neural networks for semantic segmentation.
\newblock In {\em ICCV}, 2019.

\end{thebibliography}
}

\end{document}